\documentclass[runningheads]{llncs}

\usepackage{eccv}

\usepackage{eccvabbrv}

\usepackage{graphicx}
\usepackage{booktabs}
\usepackage{longtable}
\usepackage{bbding} 
\usepackage{setspace}
\usepackage[utf8]{inputenc}
\usepackage{array} 
\usepackage{booktabs} 
\usepackage{multirow} 
\usepackage{listings}
\usepackage{xcolor}

\usepackage[accsupp]{axessibility}  

\usepackage{hyperref}

\usepackage{orcidlink}

\begin{document}

\title{RoboTwin: Dual-Arm Robot Benchmark with Generative Digital Twins (early version)} 

\titlerunning{RoboTwin(early version)}

\author{Yao Mu $^{*\dag}$\inst{1, 3}\thanks{Equal Contributions. \dag 
 Corresponding authors: Ping Luo (pluo.lhi@gmail.com) and Yao Mu (muyao@connect.hku.hk). Tianxing Chen completed this work while he was an intern at the University of Hong Kong.}  \and
Tianxing Chen $^{*}$\inst{1, 3, 4} \and 
Shijia Peng $^{*}$\inst{2, 4} \and Zanxin Chen $^{*}$\inst{2, 4} \and Zeyu Gao \inst{5} \and Yude Zou \inst{4} \and Lunkai Lin \inst{2} \and Zhiqiang Xie \inst{2} \and Ping Luo $^{\dag}$\inst{1}}


\authorrunning{Yao Mu, Tianxing Chen, et al.} 

\institute{The University of Hong Kong \and AgileX Robotics \and Shanghai AI Laboratory \and Shenzhen University \and Institute of Automation, Chinese Academy of Sciences \\ \url{https://robotwin-benchmark.github.io/early-version}}

\maketitle

\begin{abstract}
Effective collaboration of dual-arm robots and their tool use capabilities are increasingly important areas in the advancement of robotics. These skills play a significant role in expanding robots' ability to operate in diverse real-world environments. However, progress is impeded by the scarcity of specialized training data.
This paper introduces RoboTwin, a novel benchmark dataset combining real-world teleoperated data with synthetic data from digital twins, designed for dual-arm robotic scenarios. Using the COBOT Magic platform, we have collected diverse data on tool usage and human-robot interaction. We present a  innovative approach to creating digital twins using AI-generated content, transforming 2D images into detailed 3D models. Furthermore, we utilize large language models to generate expert-level training data and task-specific pose sequences oriented toward functionality. 
Our key contributions are: 1) the RoboTwin benchmark dataset, 2) an efficient real-to-simulation pipeline, and 3) the use of language models for automatic expert-level data generation. These advancements are designed to address the shortage of robotic training data, potentially accelerating the development of more capable and versatile robotic systems for a wide range of real-world applications.


\end{abstract}

\section{Introduction}
In the fast-evolving robotics field, the integration of dual-arm coordination and advanced tool use is crucial for developing sophisticated autonomous systems. These capabilities are essential for enabling robots to function effectively in diverse real-world settings such as manufacturing plants, healthcare centers, and homes. By using tools, robots can significantly expand their operational scope, adapting to a variety of tasks and challenges with greater flexibility. 
However, the advancement in these areas is substantially hindered by the lack of specialized, high-quality training data. These activities, which often require tailored solutions, are difficult to standardize and are typically not well-represented in conventional datasets.

Addressing this critical gap, we introduce "RoboTwin", a comprehensive benchmark that includes both real-world teleoperated data and corresponding synthetic data generated by a digital twin. Specifically designed for scenarios involving dual-arm robotic tool use and human-robot interactions. RoboTwin features high-quality annotations and a diversity of examples to ensure robust training and evaluation. To collect real-world data, we employ the open-source COBOT Magic platform developed by AgileX Robotics. This platform is outfitted with four AgileX Arms and four Intel Realsense D-435 RGBD cameras, mounted on a robust Tracer chassis. The data encompasses a variety of typical tasks, including tool usage and human-robot interaction.

\begin{figure}[t] 
    \centering
    \includegraphics[width=1.0\textwidth]{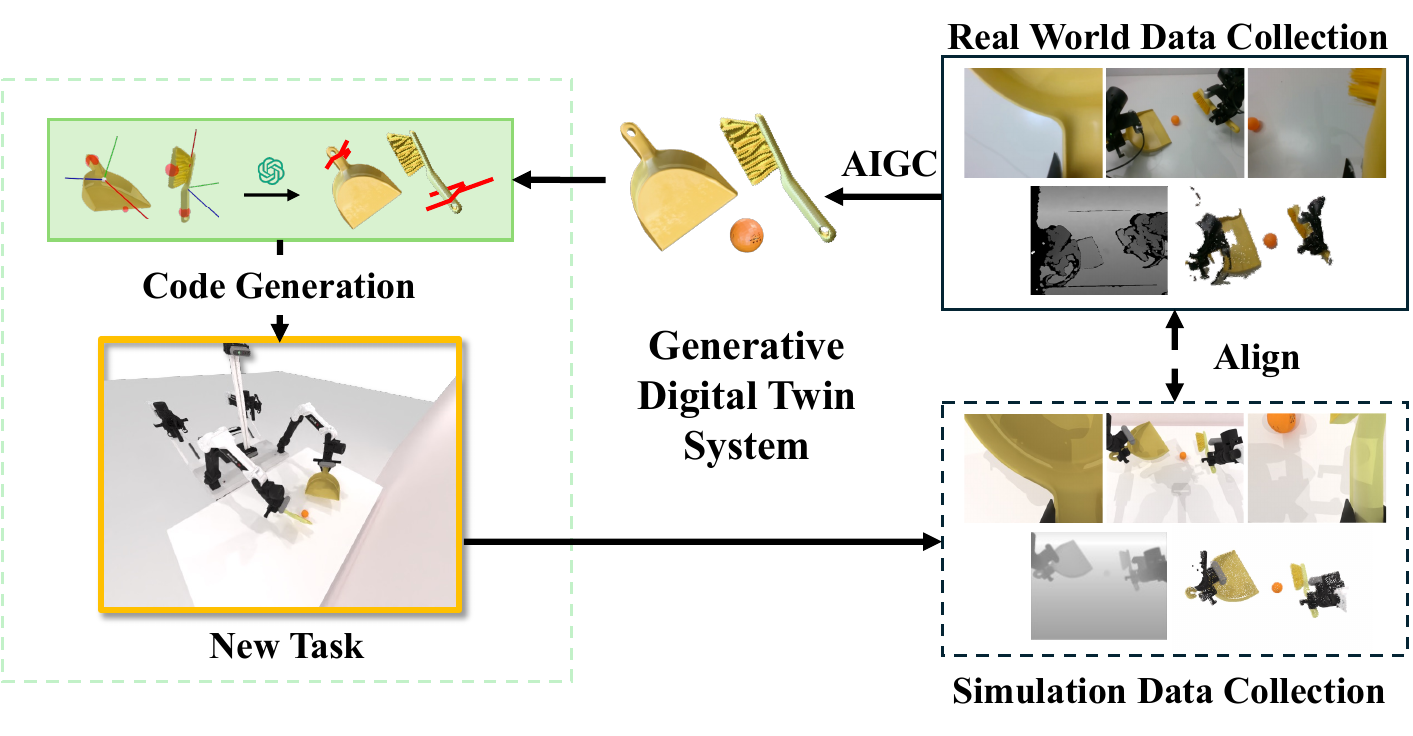}
    \caption{RoboTwin Benchmark.}
    \label{fig:hardware} 
\end{figure}

Transforming from the collection of real-world data to its virtual replication, the challenge is to create accurate and cost-effective digital twins. Traditional methodologies often rely on expensive, high-fidelity sensors, limiting their widespread adoption. To circumvent these limitations, we have developed a novel, cost-effective approach using Artificial Intelligence Generated Content (AIGC) to construct 3D models from a single 2D RGB image. This method significantly reduces costs while providing lifelike visual representations and supporting physical simulations. Our process begins with converting a 2D image into a detailed 3D model, featuring complex geometry, surface textures, and accurate details, which are crucial for realistic visualizations and simulations. We further enhance the model by defining functional coordinate axes on the object parts, enabling the automated computation of grasp poses essential for robotic manipulations.
To enhance the utility and relevance of our dataset, we have also established an innovative pipeline leveraging large language models (LLMs) for the automatic generation of expert-level training data. This methodology not only enriches the dataset with high-quality, scenario-specific examples but also integrates the versatility of LLMs in synthesizing complex interactive sequences. Integrating the reasoning capabilities of GPT4-V~\cite{2023GPT4VisionSC}, we automate the generation of task-specific pose sequences, thereby increasing the precision of task executions. Moreover, we employ GPT4-generated scripts to activate trajectory planning tools, which streamline the programming efforts and expedite the deployment of robotic systems in various environments.

The core contributions of this work are: 1) The development of "RoboTwin", a benchmark that includes both real-world teleoperated data and high-fidelity synthetic data generated for corresponding scenarios. 2) The establishment of a convenient real-to-simulation pipeline that requires only a single RGB image from the real world to generate the 3D models of target objects and corresponding scenes. 3) The utilization of large language models (LLMs) combined with simulation environment information to generate code that automatically creates expert-level data.  These advancements collectively aim to bridge the gap in robotic training data, significantly enhancing the potential for robots to learn and operate using tools in a manner that mimics human dexterity and interaction finesse.

\section{Related Work}

\subsection{Datasets and Benchmarks for Robotics}
To enhance the collection of effective demonstrations for robotic tasks, human teleoperation has traditionally been employed. In this method, a human operator manually guides a robot through various tasks~\cite{zhang2017deep, mandlekar2018roboturk, mandlekar2019scaling, mandlekar2020hitl, ebert2021bridge, jang2022bcz}. Recent advancements have extended this methodology by employing teams of human operators over prolonged periods to assemble substantial real-world datasets~\cite{ebert2021bridge, ahn2022can, jang2022bcz, rt12022arxiv}. An alternative method involves the use of algorithmic trajectory generators within simulations~\cite{james2020rlbench, zeng2020transporter, jiang2023vima, gu2023maniskill2, dalal2023imitating}, which, while efficient, often depend on privileged information and hand-designed heuristics, making them labor-intensive for arbitrary tasks.
However, current systems often fail to produce high-fidelity expert simulation data that accurately mimics data from actual machine operations. Although initiatives like MimicGen\cite{mandlekar2023mimicgen} and RoboCaca\cite{nasiriany2024robocasa} strive to generate simulated expert data using limited human demonstrations, they still heavily rely on predefined scenes and interactive objects.
To overcome these limitations, we introduce RoboTwin. This innovative system not only generates expert data and simulation scenes derived from real-world scenarios but also utilizes large language models (LLMs) to generate demonstration codes and expert data for similar tasks involving the same class of objects. This strategy significantly reduces the dependence on continuous human intervention, thereby streamlining the generation of reliable training data for robotic tasks.

\subsection{Robot Manipulation Learning Methods}
The adoption of human demonstrations to instruct robots in manipulation skills is a prevalent method in Robot Manipulation Learning \cite{sharma2018multiple,bahl2023affordances,jang2022bc,lynch2023interactive,brohan2022rt}. Among the techniques, Behavioral Cloning stands out for learning policies offline from these demonstrations. It replicates observed actions from a curated dataset \cite{pomerleau1989alvinn,zhang2017deep, mandlekar2020learning, ebert2021bridge, rt12022arxiv, jang2022bcz, dalal2023imitating, jiang2023vima}. Conversely, Offline Reinforcement Learning enhances policy learning by optimizing actions based on a predefined reward function and exploiting large datasets \cite{levine2020offline,kalashnikov2021mt, chebotar2023q, gurtler2023benchmarking, kumar2022pre, kumar2021workflow}. The Action Chunking with Transformers (ACT) technique integrates a Transformer-based visuomotor policy with a conditional variational autoencoder to structure the learning of action sequences \cite{zhao2023learning,vaswani2017transformer,sohn2015learning}. Recently, the Diffusion Policy method has gained prominence. It employs a conditional denoising diffusion process for visuomotor policy representation, effectively reducing the accumulative error in trajectory generation that is often observed in Transformer-based visuomotor policies \cite{chi2023diffusion}. The 3D Diffusion Policy \cite{ze20243d} uses point clouds for environmental observations, enhancing spatial information utilization and managing various robotic tasks in both simulated and real environments with only a small number of demonstrations.

\section{Real-to-sim transfer of the scene}

\subsection{Generative Digital Twin System}
\begin{figure}[t]
    \centering 
    \includegraphics[width=1.0\textwidth]{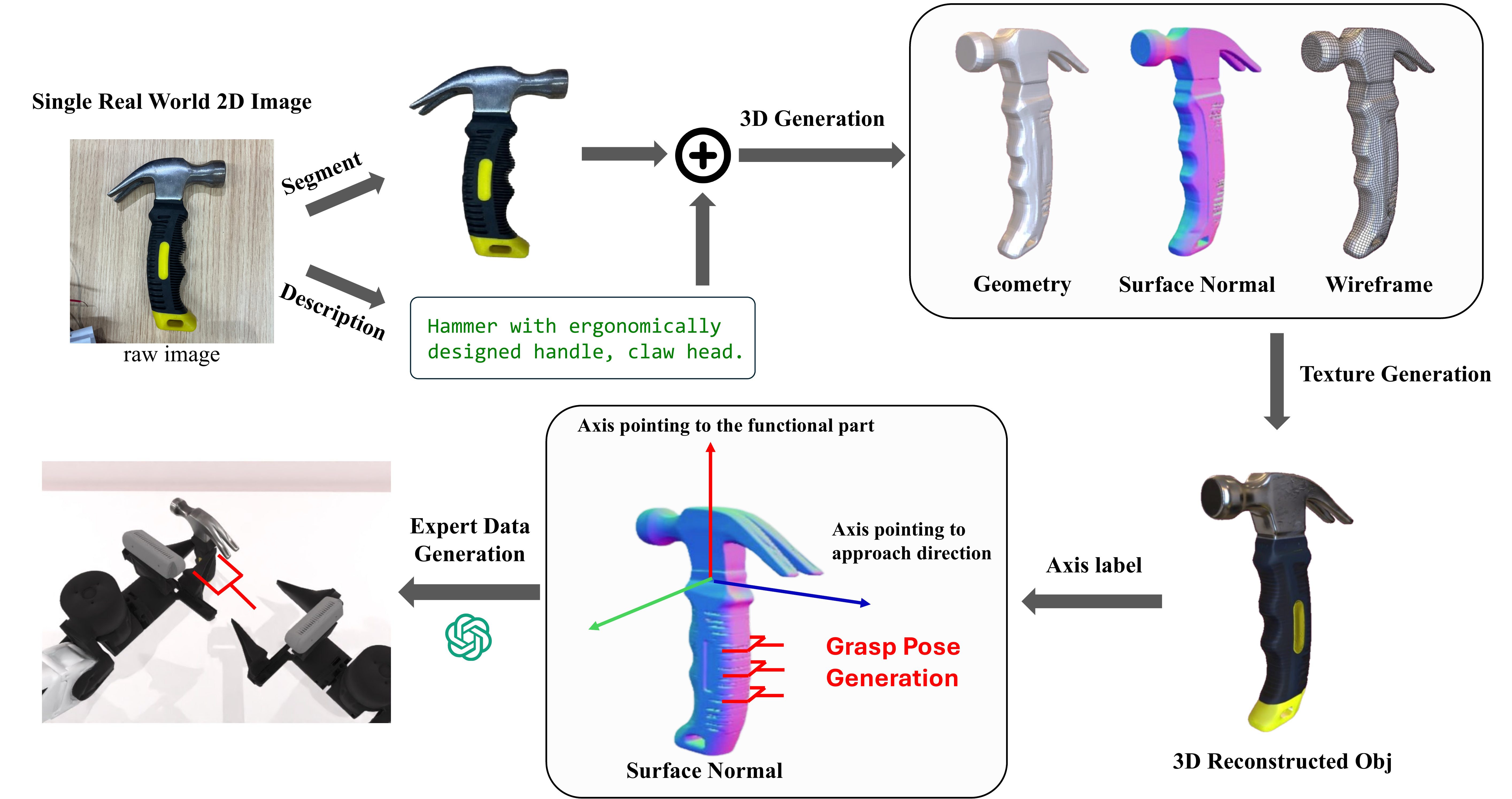}
    \caption{AIGC \& Expert Data Generation pipeline. Automatic extraction of object segmentation and textual description from a single RGB photo, followed by the generation of 3D geometry, surface normals, Wireframe, and texture maps to create a high-fidelity simulation object. With the object's surface normal and pose information, we can decompose and generate grasping postures, and leverage the large model's capabilities to zero-shot generate expert data for tasks.} 
    \label{fig:aigc} 
\end{figure}

To synthesize high-fidelity data through simulation, a major challenge is the creation of accurate and cost-effective digital twins. Traditional methods often depend on costly high-precision sensors, which can hinder widespread adoption. In response, we have developed a more economical approach using Artificial Intelligence Generated Content (AIGC) to construct 3D models from simple 2D RGB images powered by Deemos’s Rodin platform\footnote{We use Deemos's 3D digital assert  Generation Model (from text or image) Rodin: \href{https://hyperhuman.deemos.com/rodin}{https://hyperhuman.deemos.com/rodin}}. This technique significantly reduces the reliance on expensive sensors while achieving realistic visual effects and supporting physical simulations.
Our innovative pipeline commences with generating a detailed 3D mesh and texture of the target object involved in a robot's task, created from a single real-world image. This capability ensures a high-fidelity recreation of real-world scenarios within a simulated environment. The process begins by transforming a single 2D image into a 3D model that encompasses detailed geometry, surface normals, wireframes, and textures. These features enhance the visual realism and ensure compatibility with physics engines for simulations.
Once the 3D model is ready, we assign specific coordinate axes to functional parts of objects within the model. For instance, as shown in Fig. \ref{fig:aigc}, for a hammer, one axis is aligned with the hammerhead—identifying the functional part—while another axis indicates the approach direction. This strategic alignment is crucial for automating the calculation of grasp poses, which are essential for robotic manipulation and tool usage. Grasp poses are computed perpendicular to the surface normal of the functional part along the designated approach direction axis, facilitating correct and efficient tool use with minimal manual intervention.


\subsection{Expert Data Generation}
\begin{figure}[t]
    \centering 
    \includegraphics[width=1.0\textwidth]{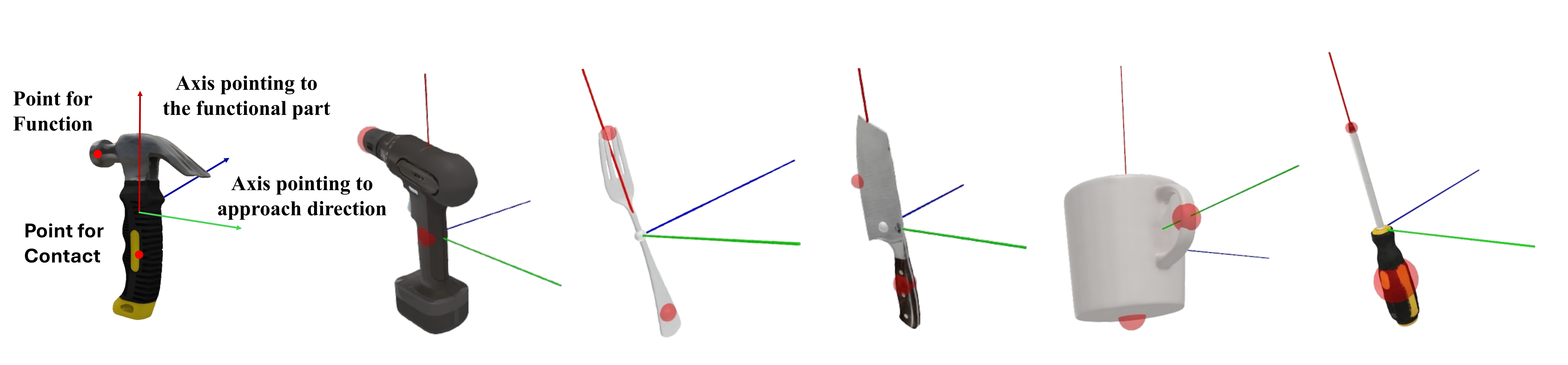}
    \caption{Point for Function and Contact, Axis pointing to the functional part and approach direction} 
    \label{fig:aigc} 
\end{figure}

We leverage the reasoning capabilities of GPT4-V~\cite{2023GPT4VisionSC} to write code that calculates the relationships between key poses and the functional coordinate axes of objects. GPT4-V analyzes task requirements and generates a sequence of poses that align with these requirements, ensuring precise task execution. We also generate code via GPT4~\cite{achiam2023gpt} to invoke trajectory planning tools based on the computed poses. This automation substantially decreases the time and labor associated with manual programming, facilitating the swift deployment of robotic systems across diverse applications. It also offers a scalable approach for generating high-quality data essential for robotic learning.

\begin{figure}[htbp] 
    \centering    \includegraphics[width=0.99\textwidth]{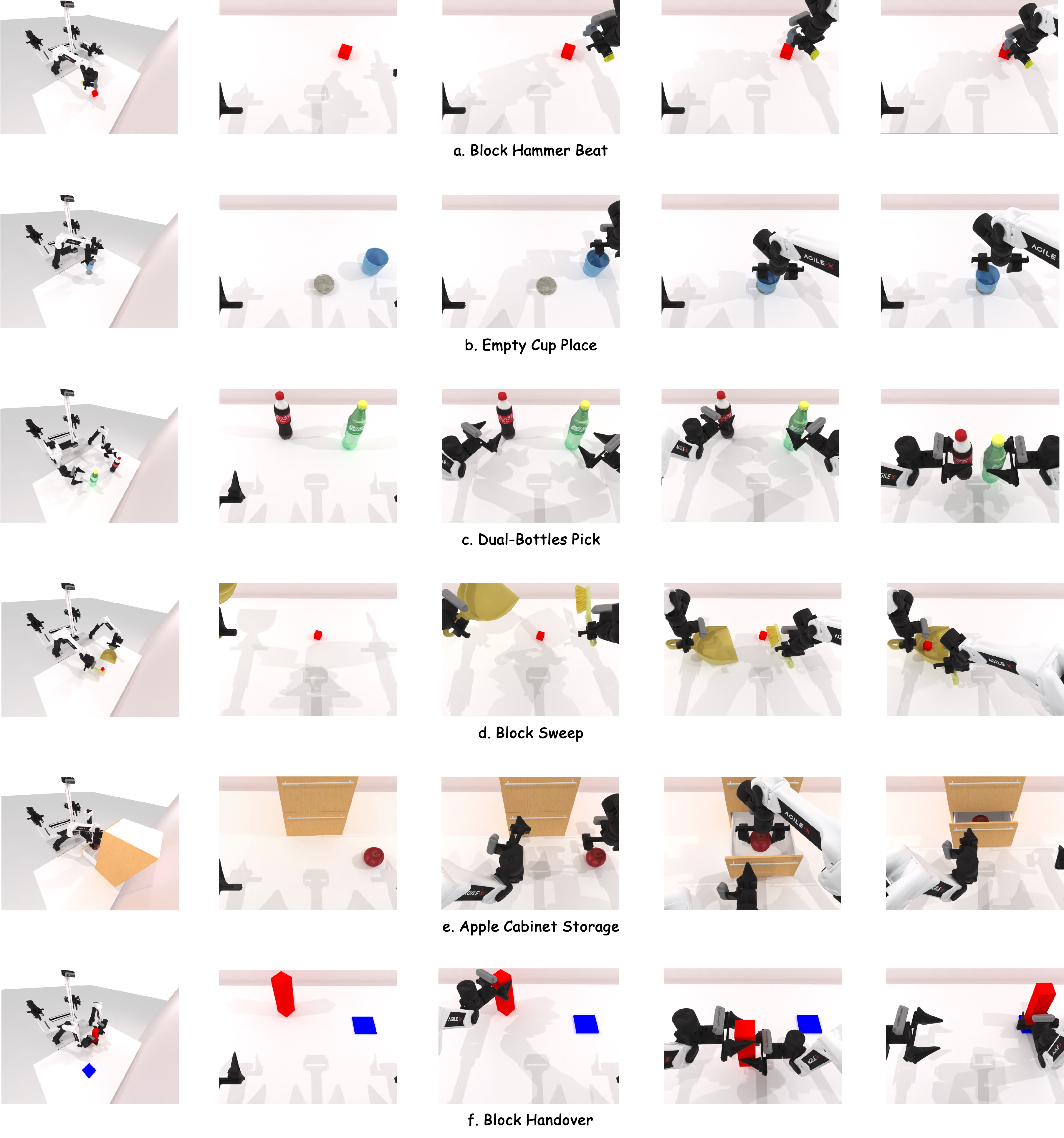}
    \caption{Task Execution of RoboTwin Benchmark.}
    \label{fig:benchmark_demo} 
\end{figure}
\section{Benchmark}
To further research and development in this area, as shown in Fig. \ref{fig:benchmark_demo}, we introduce a comprehensive benchmark specifically designed to assess dual-arm robots in a variety of scenarios. This benchmark encompasses a diverse set of tasks, each presenting unique challenges that are critical for assessing the dexterity, coordination, and operational efficiency of robotic arms in a simulated environment. The tasks range from simple object manipulation to complex, coordinated actions requiring synchronized movements of both arms. Appendix A.2 outlines the specific tasks and their descriptions, providing a clear framework for comparative analysis and further development of advanced robotic capabilities. 
For each task, we provide a robust API that supports the generation of expert data across infinitely variable scenarios, such as different object placements and environmental conditions. This feature allows researchers to extensively test and refine the adaptability and precision of robotic systems under controlled yet varied conditions. Additionally, an offline dataset is available for each task, offering pre-generated expert data to facilitate offline training and algorithm benchmarking. This benchmark aims to bridge the gap between theoretical robotic control models and their practical implementation, ensuring that robotic systems can perform reliably in dynamic, real-world environments.




\section{Real-world Dataset}
\begin{figure}[htbp] 
    \centering
    \includegraphics[width=1.0\textwidth]{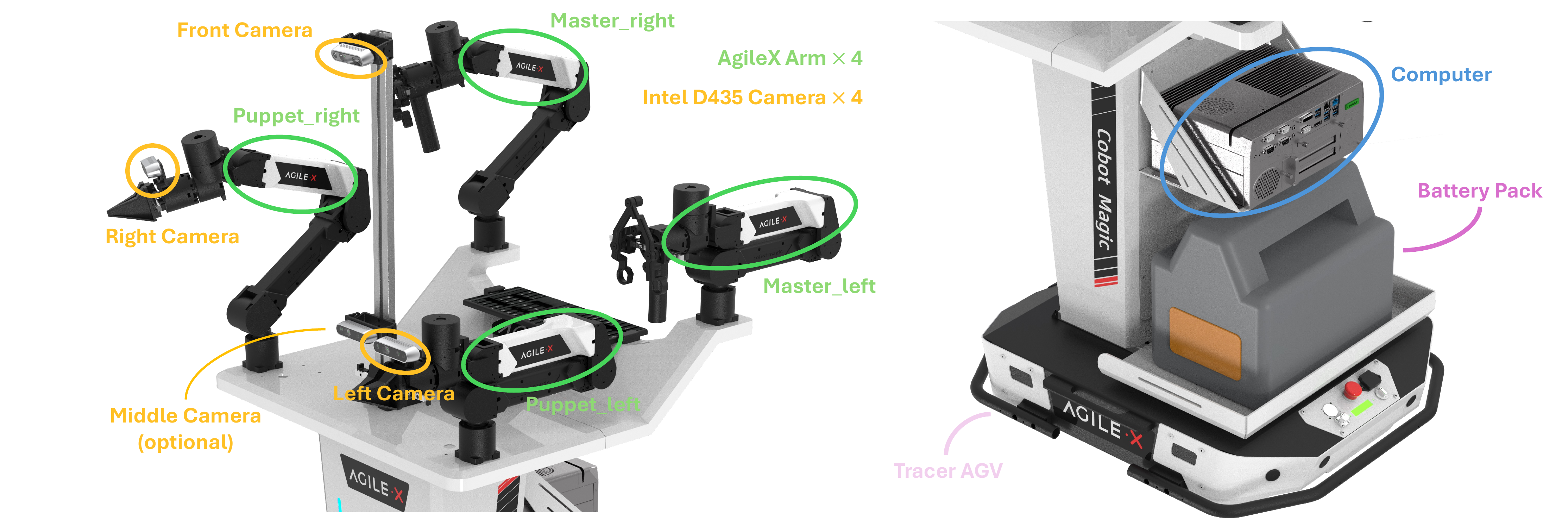}
    \caption{Illustration of our robot platform, with the capabilities for teleoperation, mobility, and data acquisition.}
    \label{fig:hardware} 
\end{figure}

For the acquisition of real-world data, we employed the  Cobot Magic \footnote{Platform Introduction: \href{https://global.agilex.ai/products/cobot-magic}{https://global.agilex.ai/products/cobot-magic}} platform from AgileX Robotics, which is equipped with four AgileX Arms and four Intel Realsense D-435 RGBD cameras and is built on the Tracer chassis. These cameras are strategically positioned: one on the high part of the stand for an expansive field of view, two on the wrists of the robot's arms, and one on the low part of the stand which is optional for use. The front, left, and right cameras capture data simultaneously at a frequency of 30Hz, as depicted in Figure \ref{fig:hardware}. The data collection and alignment are facilitated by tools provided by the ARIO Data Alliance, available at our GitHub repository\footnote{Tool for data alignment: \href{https://github.com/ario-dataset/ario-tools}{https://github.com/ario-dataset/ario-tools}}. Each captured frame consists of three images from the cameras, each providing an RGB and depth image at a resolution of 640$\times$480 pixels. Additionally, the data includes the poses of the robotic arms' joints and end-effectors for both master and slave configurations, encompassing both left and right arms. All data storage and formatting adhere to the unified standards established by the ARIO Data Alliance.


Our dataset task design features two major highlights: a focus on human-robot interaction and tool usage. As shown in Appendix A.3, we have designed 17 tasks, 9 of which emphasize tool usage, 5 involve interpersonal interactions and 6 tasks are dual-arm. We collected 30 trajectories for each task. During trajectory collection, we broke down the tasks into multiple stages and conducted slower data collection for key sub-trajectories that required precise operations, enhancing the detail of the trajectories for better model learning.

\section{Experiment}
Our experimental aim is not to delve into the design choices of different strategy networks but to explore the correctness and effectiveness of our Benchmark expert data. Our experiments are intended to verify: a) the rationality of the COBOT Magic platform settings, and b) the effectiveness of the automatically generated expert data.

We utilized the 3D Diffusion Policy (DP3) \cite{ze20243d} to test six tasks within the benchmark, with each task being tested using strategies trained from 10 sets, 20 sets, and 50 sets of expert data, respectively, to obtain the success rates. 
%
\begin{table}[t]
    \footnotesize
  \centering
        \caption{The testing results for DP3 on various tasks, trained with varying amounts of expert data.}
    \begin{tabular}{cccc}
    \toprule
          \textbf{Task} & \textbf{\hspace{1em}10 demo.\hspace{1em}} & \textbf{\hspace{1em}20 demo.\hspace{1em}} & \textbf{\hspace{1em}50 demo.\hspace{1em}} \\
    \midrule
        Block Hammer Beat & 24\% & 56\% & 80\% \\ 
        Empty Cup Place & 10\% & 60\% & 96\% \\ 
        Dual-Bottles Pick & 10\% & 42\% & 74\% \\ 
        Block Sweep & 28\% & 70\% & 86\% \\ 
        Apple Cabinet Storage & 30\% & 57\% & 64\%  \\
        Block Handover & 50\% & 90\% & 98\% \\
    \bottomrule
    \end{tabular}%

       \label{tab:experiment}
\end{table}
%
The experimental results, as summarized in Table~\ref{tab:experiment}, demonstrate the performance of the 3D Diffusion Policy (DP3) across six tasks, each trained with varying quantities of expert demonstration data (10, 20, and 50 sets). Notably, for the "Block Hammer Beat" task, the success rate improved from 24\% with 10 demonstrations to 80\% with 50 demonstrations. The "Empty Cup Place" task saw success rates soar from 10\% to 96\% with increased demonstrations. The "Dual-Bottles Pick" task success rates climbed from 10\% to 74\% as demonstrations increased. The "Block Sweep" task success improved steadily from 28\% to 86\%, while the "Apple Cabinet Storage" task showed more modest gains, from 30\% to 64\%. The "Block Handover" task achieved the most significant improvement, reaching 98\% success with 50 demonstrations, up from 50\%.
These results suggest a strong correlation between the number of expert demonstrations and task success, highlighting the effectiveness of the automatically generated expert data in enhancing task performance on the COBOT Magic platform. The data further underscores the importance of ample training examples in the development of robust strategies for complex tasks.

\section{Conclusion}
In this study, we introduce RoboTwin, a benchmark integrating real-world and synthetic data to evaluate dual-arm robots, addressing the significant shortage of specialized training data in robotics. Our dataset, developed using the AgileX Robotics platform and enhanced through generative digital twins powered by Deemos’s Rodin platform, effectively accelerates the training of robotic systems, enabling performance improvements across diverse tasks. Our results demonstrate the potential of this hybrid data approach to refine robotic dexterity and efficiency, providing a scalable tool that could revolutionize robotic research and applications.



\bibliographystyle{splncs04}
\bibliography{main}

\end{document}